\documentclass{article}

\usepackage{PRIMEarxiv}

\usepackage[utf8]{inputenc} 
\usepackage[T1]{fontenc}    
\usepackage{hyperref}       
\usepackage{url}            
\usepackage{booktabs}       
\usepackage{amsmath,amssymb,amsfonts}
\usepackage{nicefrac}       
\usepackage{microtype}      
\usepackage{lipsum}
\usepackage{fancyhdr}       
\usepackage{graphicx}       
\graphicspath{{media/}}     

\usepackage{multirow,tabularx}
\usepackage{placeins}
\usepackage[utf8]{inputenc} 
\usepackage[T1]{fontenc}    
\usepackage{hyperref}       
\usepackage{url}            
\usepackage{booktabs}       
\usepackage{nicefrac}       
\usepackage{ragged2e}
\usepackage{caption}
\usepackage{subcaption}
\usepackage{setspace}
\usepackage{svg}
\usepackage{float}
\usepackage[linesnumbered,ruled,noline]{algorithm2e}
\usepackage{framed}
\usepackage{lineno}
\usepackage{adjustbox}
\usepackage{wrapfig}
\usepackage{lscape}
\usepackage{rotating}
\usepackage{epstopdf}
\usepackage{numprint}

\SetCommentSty{mycommfont}

\SetKwInOut{Input}{input}
\SetKwInOut{Output}{output}
\SetKw{Kwminibatch}{sample\_minibatch\_by\_index}
\SetKw{Kwduplicate}{duplicate}

\npdecimalsign{.}
\nprounddigits{3}

\def\agree{\textit{agree}}
\def\joint{\textit{joint}}

\title{SiamAF: Learning Shared Information from ECG and PPG Signals for Robust Atrial Fibrillation Detection}

\author{
  Zhicheng Guo, Cheng Ding, Duc H. Do, Amit Shah, Randall J. Lee, Xiao Hu, Cynthia Rudin
}

\begin{document}
\maketitle

\begin{abstract}
Atrial fibrillation (AF) is the most common type of cardiac arrhythmia. It is associated with an increased risk of stroke, heart failure, and other cardiovascular complications, but can be clinically silent. Passive AF monitoring with wearables may help reduce adverse clinical outcomes related to AF.  Detecting AF in noisy wearable data poses a significant challenge, leading to the emergence of various deep learning techniques. Previous deep learning models learn from a single modality – either electrocardiogram (ECG) or photoplethysmography (PPG) signals. However, deep learning models often struggle to learn generalizable features and rely on features that are more susceptible to corruption from noise, leading to sub-optimal performances in certain scenarios, especially with low-quality signals. Given the increasing availability of ECG and PPG signal pairs from wearables and bedside monitors, we propose a new approach, SiamAF, leveraging a novel Siamese network architecture and joint learning loss function to learn shared information from both ECG and PPG signals. At inference time, the proposed model is able to predict AF from either PPG or ECG and outperforms baseline methods on three external test sets. It learns medically relevant features as a result of our novel architecture design. The proposed model also achieves comparable performance to traditional learning regimes while requiring much fewer training labels, providing a potential approach to reduce future reliance on manual labeling.
\end{abstract}


\section{Introduction}

Atrial fibrillation (AF) is the most prevalent form of cardiac arrhythmia affecting approximately 1-2\% of the general population and up to 9\% of individuals aged 65 years or older. \cite{afperc, afperc2} AF contributed to over 183K deaths in the US alone in 2019 \cite{cdcaf}. It is estimated to be present in 6-12 million people worldwide, and AF prevalence is projected to continuously increase in the next 30 years \cite{afibbad}. Paroxysmal AF episodes often present little to no symptoms -- that patients do not usually notice immediately -- that are crucial precursors to more serious health conditions, including ischemic stroke and congestive heart failure. Therefore, the detection of AF episodes has become imperative in the treatment and prevention of cardiovascular disease. AF detection has been transformed by the growing popularity of wearable devices with photoplethysmography (PPG) sensors, such as smartwatches. These devices are widely available, providing a non-invasive means for continuous heart rhythm monitoring in day-to-day ambulatory settings. The availability of smartwatch PPG data, as well as electrocardiogram (ECG) and PPG monitoring in hospital settings, has created an unprecedented surge in heart monitoring data availability. The sheer volume of data produced from wearable devices and bedside monitors has rendered manual reviewing of recorded signals intractable, leading to a growing demand for automated or assistive systems that can help clinicians detect AF. Therefore, there is a pressing demand for the development of algorithms capable of timely and precise detection of AF. 

There have been numerous works leveraging machine learning \cite{AFML1, AFML2, AFML3, AFML5, AFML6, AFML7, AFML8} and specifically deep learning algorithms \cite{AFCNN1, AFCNN2, AFCNN3, AFCNN4, AFCNN5, AFCNN6, stanford} for automating AF detection. These models were trained either using ECG or PPG signals, despite the availability of both modalities (PPG signals are usually labeled using ECG as the gold standard). Both ECG and PPG signals carry crucial information for detecting AF. Clinically, ECG signals reveal different types of arrhythmias that may present similarly in PPG; the addition of ECG signal features to PPG AF detection pipelines should thus improve detection performance. Moreover, if one of the signal modalities contains artifacts (e.g., motion artifacts, environmental lighting artifacts), the other can compensate for the loss of information. Although ECG and PPG use different collection mechanisms (i.e., ECG uses electrodes, PPG uses photodetectors) to measure the pulsatile changes, heart rate variability and other factors can be monitored by both signals. Therefore, we hypothesize it will be beneficial to leverage both signal modalities for AF detection. 

In this work, we introduce a novel AF detection algorithm, SiamAF, which learns robust and medically relevant features from both ECG and PPG signals. As we will demonstrate, the model successfully learns the shared information between the two signal modalities through a novel model architecture and loss function.

This study has the following contributions:
\begin{itemize}
    \item To our knowledge, this is the first study leveraging shared information between ECG and PPG signals for AF detection. Our novel architecture and loss function design encourage the model to learn shared information and improve prediction robustness. 
    \item The proposed method outperforms baseline methods, including deep mutual learning \cite{deepmut} and previous CNN-based ECG or PPG single modality AF detection networks \cite{stanford} on three external test sets containing diverse patient conditions and recording hardware.
    \item We investigate the information learned by the model and verify the effectiveness of our proposed method through dimension reduction and visualizations of latent features.
    \item Using only 1\% of the training labels, the proposed method outperforms baseline methods. The proposed model learns signal features by learning shared information between ECG and PPG signals in a contrastive fashion in addition to learning to classify AF. The semi-supervised learning setup utilizes data more efficiently compared to previous traditional models learning from a single modality.
    \item The proposed method learns a single model that can be used for AF detection on both ECG or PPG signals with no performance sacrifice for either modality.
\end{itemize}

\section{Related Works}
We consider related work in AF detection with hand-crafted features, deep learning for AF detection, and Siamese networks.

\textbf{Hand-crafted features for AF detection.}
There have been multiple past works that developed hand-crafted features of the ECG or PPG input signals for training machine learning-based AF classifiers. These include the Root Mean Square of the Successive Difference of peak-to-peak intervals (RMSSD), Shannon entropy (ShE), spectral analysis, dynamic time warping for waveform shape analysis and template matching, as well as other statistical features such as mean, variance, standard deviation, skewness and kurtosis of input signals \cite{AFML1, AFML2, AFML3, AFML5, AFML6, AFML7, AFML8}. These features are fed into standard machine learning classification algorithms. 

Most hand-crafted features rely on accurate peak detection in the raw signals, which is often unreliable, given poor signal qualities. These challenges have made the hand-crafted features approach difficult for widespread applications.

\textbf{Deep learning for AF detection.}
Deep learning networks automatically learn and extract features from raw inputs. Deep convolutional neural networks (DCNN/CNN) have been popular as feature extractors for both images \cite{medicalCV} and time series due to their unique ability to preserve useful information. There have been several CNN-based AF detection algorithms \cite{AFCNN1, AFCNN2, AFCNN3, AFCNN4, AFCNN5, AFCNN6, stanford}. Some researchers convert the 1-D signals into 2-D images using the short-term Fourier transform (STFT) and stationary wavelet transform (SWT) \cite{AFCNN2D1, AFCNN2D2} for 2D-CNNs. Due to the time series nature of ECG and PPG signals, many researchers leverage recurrent neural networks (RNNs) for AF detection due to their ability to capture temporal relations and their flexibility with variable length inputs. Some of its variants, including the long short-term memory network (LSTM) and recurrent-convolutional neural network, have also been used for AF detection \cite{AFRNN1, AFRNN2, AFRNN3, AFRNN4, AFRNN5}. 
While deep neural networks are more adaptable than hand-crafted features-based machine learning methods, they require significantly more labeled data to achieve performance equivalent to the best AF detectors. Deep learning models are prone to overfitting, especially on trivial task-irrelevant features. This is often demonstrated through the sensitivity of deep neural nets to signals that are low quality and contain signal artifacts; we can expect worse performances of the trained models on these low-quality signals \cite{overfit, cmc}. Deep learning methods also suffer from a lack of interpretability. It is difficult, if not impossible, for humans to understand the discovered features and decision processes of the deep neural networks; combined with the susceptibility to overfitting, it leads to an increased risk of undetected failure in real-world applications when operating conditions change or the model fails to generalize \cite{chestxray, eyedeployment}. 

\textbf{Two Modalities.}
Previous studies only trained their models using either ECG or PPG signals \cite{AFML1, AFML2, AFML3, AFML5, AFML6, AFML7, AFML8, AFCNN1, AFCNN2, AFCNN3, AFCNN4, AFCNN5, AFCNN6, stanford, AFDL6, AFDL7, AFDL8, AFDL9, AFRNN1, AFRNN2, AFRNN3, AFRNN4, AFRNN5, AFCNN2D1, AFCNN2D2, AFDL1MLP}, despite the availability of both modalities measured simultaneously on the same individuals. As we will show, both ECG and PPG signals carry crucial information for detecting AF; neither modality should be wasted during training. 
 
\textbf{Siamese networks.}
The Siamese network architecture (models with two branches or subnetworks) is useful for dual input scenarios. In recent years, Siamese networks have seen a rise in popularity due to their usefulness for self-supervised learning \cite{simclr, byol, simsiam}. However, Siamese self-supervised networks are difficult to train and rely heavily on image augmentations.

\section{Methods}
\label{sec:method}
\begin{figure*}[hbt!]
  \centering
     \includegraphics[width=.8\textwidth]{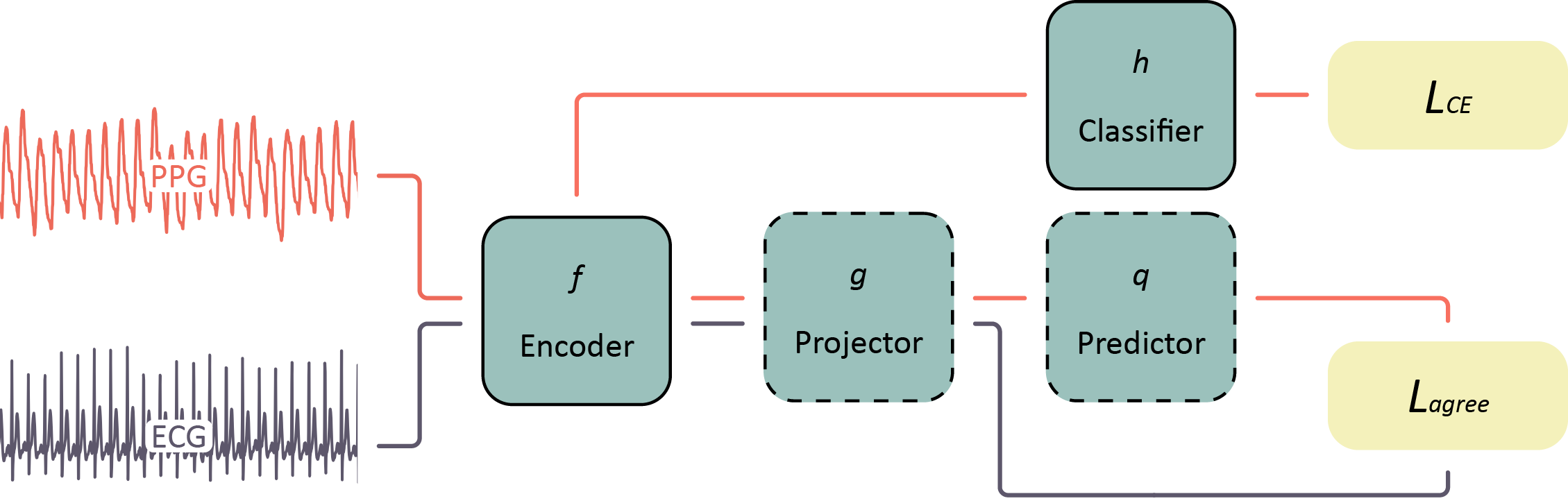}
     \caption{Here we show the architecture of our proposed framework for learning from PPG and ECG and predicting (i.e., testing) on only either ECG or PPG. At training time, the model takes both ECG and PPG signals as inputs. In each training iteration, the two signal modalities take turns flowing through the network following each of the colored paths; i.e., the PPG signal flows through the red path and the ECG signal flows through the purple path, then the ECG signal flows through the red path and the PPG signal flows through the purple path. In the configuration shown in the figure, the PPG and ECG signal inputs will go through the encoder $f$ and the projector $g$. The learned features of the PPG signal will pass through the predictor $q$ to map to the latent space of ECG features. We optimize an agreement loss between the predicted latent feature vector of the PPG signal and the projected latent feature vector of the ECG signal; we also optimize a supervised cross-entropy loss for the output of the classifier $h$ (which takes PPG latent features as inputs). For a detailed description of the training process, please see Section \ref{sec:method} and Algorithm \ref{alg:algo_1}. After the training is complete, we save only the encoder and the classifier for future predictions.}
     \label{fig:arch}
\end{figure*} 

The training data is denoted as $D=\{(x^{ECG}_{i}, x^{PPG}_{i}, y_i)\}^{|D|}_{i=1}$. Our joint training framework takes a pair of time-synchronized ECG and PPG signals as input ($x^{ECG}_{i}$ and $x^{PPG}_{i}$), and a binary label of AF/non-AF ($y_i$) generated automatically from the hospital alarm. The auto-generated labels contain a certain amount of noise. Our proposed framework learns by simultaneously maximizing the agreement between the latent projections of the ECG-PPG signal pairs, and minimizing the misclassification error of the classifications from the ECG and PPG signals. Our framework is a Siamese network with five major components:
\begin{itemize}
    \item  A neural network encoder $f(\cdot)$ which takes both the ECG $x^{ECG}_{i}$ and PPG $x^{PPG}_{i}$ signals as inputs. The encoder extracts feature representations from the raw inputs. In theory, the architecture of the encoder can take any form, so future applications can choose a different architecture than the specific one used here. Here we use a 1-D version of the ResNet-34 \cite{resnet} architecture for the encoder due to its effectiveness for physiological signals.
    \item A multi-layer-perception projector $g(\cdot)$ with one hidden layer. The projector takes the feature outputs from the encoder as input.  The projector will map the extracted features from the encoder to a latent space where we maximize the agreements between the features of ECG and PPG signals. The projected views are denoted as $z^{ECG}_i = g(f(x^{ECG}_{i}))$ for an ECG input and $z^{PPG}_i = g(f(x^{ECG}_{i}))$ for a PPG input. (Function $g$ is the same regardless of whether we translate ECG to PPG or vice versa.)
    \item A multi-layer-perception predictor $q(\cdot)$ with no hidden layers. The predictor takes one of the projector's outputs from the ECG-PPG pair, and further maps the projected views. We denote the predictor's predictions as  $q(z^{ECG}_i)$ for an ECG input and $q(z^{PPG}_i)$ for a PPG input. (Here the functions $f$, $g$, and $q$ are the same regardless of whether we translate ECG to PPG or vice versa and regardless of whether we predict for ECG or PPG.)
    \item A supervised branch classification head $h(\cdot)$, consists of one linear layer. It takes the output from the encoder $f(\cdot)$ as input, and produces logits of $p^{ECG}_i = h(f(x^{ECG}_{i}))$ for an ECG input and $p^{PPG}_i = h(f(x^{PPG}_{i}))$ for a PPG input.
    \item A loss function combing both the agreement and classification objectives. The agreement objective is to maximize the cosine distance between the $(q(z^{ECG}_i),z^{PPG}_i)$ pair and between the $(q(z^{PPG}_i),z^{ECG}_i)$ pair. We formulate the agreement objective loss function as follows:
    \begin{equation}
    \begin{split}
    & \mathcal{L}_{\agree}(x^{PPG}_i, x^{ECG}_i) \\
     & =  - \frac{\left\langle q(z^{ECG}_i), z^{PPG}_i\right\rangle}{\left\|q(z^{ECG}_i)\right\|_{2} \cdot\left\|z^{PPG}_i\right\|_{2}} - \frac{\left\langle q(z^{PPG}_i), z^{ECG}_i\right\rangle}{\left\|q(z^{PPG}_i)\right\|_{2} \cdot\left\|z^{ECG}_i\right\|_{2}}.
    \end{split}
    \label{eq:agree}
    \end{equation}
    Here $\left\|\cdot\right\|_{2}$ is the $\ell_2$-norm and $\left\langle\cdot\right\rangle$ is the inner product. To minimize the misclassification error, we employ the cross-entropy loss. By using a hyper-parameter $\lambda$, the final loss function is formulated as a weighted combination of the agreement loss and the classification loss; it is defined as follows:
    \begin{equation}
        \begin{aligned}
            & \mathcal{L}_{\joint}(x^{PPG}_i, x^{ECG}_i, y_i) \\ 
            & = \mathcal{L}_{\agree}(x^{PPG}_i, x^{ECG}_i) \\
            & + \lambda \cdot (\mathcal{L}_{\textit{CE}}(p^{PPG}_i, y_i)+\mathcal{L}_{\textit{CE}} (p^{ECG}_i, y_i)).
        \end{aligned}
        \label{eq:joint}
    \end{equation}
\end{itemize}
In cases of incomplete ECG and PPG pairs (either ECG or PPG signals) in the training data, we can accommodate that in our framework by adding two additional loss terms, one for additional PPG data and one for additional ECG data:
   \begin{equation}
        \begin{aligned}
            & \mathcal{L}_{\joint}(x^{PPG}_i, x^{ECG}_i, y_i) \\ 
            & = \mathcal{L}_{\agree}(x^{PPG}_i, x^{ECG}_i) \\
            & + \lambda \cdot (\mathcal{L}_{\textit{CE}}(p^{PPG}_i, y_i)+\mathcal{L}_{\textit{CE}} (p^{ECG}_i, y_i)) \\
            & +\lambda_{1} \cdot (\mathcal{L}_{\textit{CE}}(p^{PPG_{\textit{unpaired}}}_i, y_i)+\mathcal{L}_{\textit{CE}} (p^{ECG_{\textit{unpaired}}}_i, y_i)).
        \end{aligned}
        \label{eq:joint_w_unpaired}
    \end{equation}

In addition to the previously introduced model information, Algorithm \ref{alg:algo_1} illustrates the training algorithm of the proposed framework. After the training is complete, we discard the projector $g$ and predictor $q$ and keep the learned encoder $f$ and the classification head $h$ for future predictions. At test time, the learned network can work with either ECG or PPG signals without requiring an ECG-PPG pair.

\begin{algorithm}[hbt!]

    \setstretch{1.35}
    \DontPrintSemicolon
    \caption{Joint training algorithm}
    \label{alg:algo_1}
    \Input{$D=\{(x^{ECG}_{i}, x^{PPG}_{i}, y_i)\}^{|D|}_{i=1}$, network modules for $f, g, q, h$, hyper-parameter $\lambda$}
    
     \For{\Kwminibatch D}{
        \textcolor{blue}{// pass ECG signal}\\
        $z^{ECG}_i = g(f(x^{ECG}_{i}))$ \tcp* {ECG latent projection}
        $q(z^{ECG}_i)$ \tcp* {ECG latent prediction}
        $p^{ECG}_i = h(f(x^{ECG}_{i}))$ \tcp* {ECG class. logits}
        \textcolor{blue}{// pass PPG signal}\\
        $z^{PPG}_i = g(f(x^{PPG}_{i}))$ \tcp* {PPG latent projection}
        $q(z^{PPG}_i)$ \tcp* {PPG latent prediction}
        $p^{PPG}_i = h(f(x^{PPG}_{i}))$ \tcp* {PPG class. logits}
        Update $f, g, q, h$ to minimize $\mathcal{L}_{\joint} (x^{PPG}_i, x^{ECG}_i, y_i)$
     }
    \Return $f, h$
    
\end{algorithm}

\section{Datasets}
\label{sec:datasets}
We used a large-scale dataset from Institution A (University of California San Francisco medical center) for training and three additional datasets for evaluation, including a dataset from Institution B (University of California Los Angeles medical center), as well as the Stanford dataset test split \cite{stanford}, and the Simband dataset \cite{simband}. 

\subsection{Institution A dataset}
The Institution A (University of California San Francisco medical center) dataset contains 28539 patients in hospital settings; the patients' continuous ECG and PPG signals were recorded from the bedside monitors. The bedside monitor produced alarms for events including atrial fibrillation (AF), premature ventricular contraction (PVC), Tachy, Couplet, etc. This study focuses on AF, PVC, and normal sinus rhythm (NSR). The samples with PVC and NSR labels were combined into the Non-AF samples group, thus forming the AF vs Non-AF binary classification task. The continuous ECG and PPG signals were sliced into 30-second non-overlap segments sampled at 240Hz (each signal strip contains 7,200 timesteps). The 30-second segments were then down-sampled to 2,400 timesteps (80Hz sampling rate). During the pre-processing step, invalid samples (e.g., empty signals files, missing ECG or PPG signals) were also filtered out. There are four ECG channels in this dataset, we used the first ECG channel for our study due to its resemblance to wearable device outputs. The dataset was split into the train and validation splits by patient ids. The train split of the Institution A dataset contains 13,432 patients, 2,757,888 AF signal segments, and 3,014,334 Non-AF signal segments; the validation split contains 6,616 patients, 1,280,775 AF segments, and 1,505,119 Non-AF segments. Due to the automatic nature of bedside monitor-generated labels, the dataset likely contains label noise. For a detailed description of the alarm labeling process and the additional pre-processing steps please see our parallel work \cite{cmc}. For detailed demographic information of patients who participated in this study, please refer to Tab.\ref{tab:ucsf_demographics}. The data collection process followed strict guidelines and the study was approved by the Institutional Review Board (IRB number:14-13262). A waiver of informed patient consent was granted for this study, as the investigation solely involved the analysis of de-identified patient data.

\subsection{Institution B dataset (Test set 1)}
The Institution B (University of California Los Angeles medical center)  dataset contains 126 patients in hospital settings and simultaneous continuous ECG and PPG signals were collected at Institution B. The patients have a minimum age of 18 and a maximum age of 95 years old and were admitted from April 2010 to March 2013. The continuous signals were sliced into 30-second non-overlapping segments and again downsampled to 80Hz sampling rate with 2,400 timesteps in each signal. The dataset contains 38,910 AF and 220,740 Non-AF segments. Board-certified cardiac electrophysiologists annotated all AF episodes in the Institution B datasets. Here the PPG signals are collected from the fingertips. The use of this dataset for the study was approved by the IRB (IRB approval number: 10-000545).  A waiver of informed patient consent was obtained for this study. 

\subsection{Simband dataset (Test set 2)}
The Simband dataset \cite{simband} contains 98 patients in ambulatory settings from Emory University Hospital (EUH), Emory University Hospital Midtown (EUHM), and Grady Memorial Hospital (GMH). The patients have a minimum age of 18 years old and a maximum age of 89 years old; patients were admitted from October 2015 to March 2016. The ECG signals were collected using Holter monitors, and the PPG signals were collected from a wrist-worn Samsung Simband. The signals used for testing were 30-second segments with 2,400 timesteps after pre-processing. This dataset contains 348 AF segments and 506 Non-AF segments. The signals in this dataset were reviewed and annotated by an ECG technician, physician study coordinator, and cardiologist. The data were collected with patient consent and IRB approval (IRB approval number: 00084629).

\subsection{Stanford dataset (Test set 3)}
The Stanford dataset \cite{stanford} contains 107 AF patients, 15 paroxysmal AF patients, and 41 healthy patients. The 41 healthy patients also undergo an exercise stress test. All signals in this dataset were recorded in ambulatory settings. The ECG signals were collected from an ECG reference device, and the PPG signals were collected from a wrist-worn device. The signals were sliced into 25-second segments by the original author. In this study, we augmented the signal to 30 seconds by pasting the first 5 seconds of each signal to the end and re-sampled the signals to 2,400 timesteps. The dataset contains 52,911 AF segments and 80,620 Non-AF segments. In the evaluations, we use the test split generated by the authors of the Stanford dataset. The PPG signals in this dataset were manually annotated and reviewed by several cardiologists following computerized reference ECG signals.

\section{Experiments and Evaluation Results}
\label{sec:exp}

\subsection{Performance comparison with baseline methods}
We verify our framework's performance advantage by comparing it to three baseline methods: (1a) A ResNet-34 \cite{resnet} classifier trained on only PPG data; (1b) A ResNet-34 \cite{resnet} classifier trained on only ECG data; (2) A Deep Mutual Learning \cite{deepmut} (Deep Mut$.$ for short) network using ResNet-34 as encoders; (3) The public DeepBeat model from the authors of \cite{stanford}. The ResNet-34 baselines serve as representations of performance in previous deep learning AF detection studies, where convolutional neural networks appear as a popular choice \cite{AFCNN1, AFCNN2, AFCNN3, AFCNN4, AFCNN5, AFCNN6}. The Deep Mut$.$ baseline is known for its ability to extract shared information from dual inputs through ``mutual learning''; which could be suitable for our intended use.  The performance of the models is evaluated on the three external datasets we introduced in Section \ref{sec:datasets}. In this work, we do not compare against hand-crafted feature-based AF classifiers due to their lackluster performance compared to deep learning methods, demonstrated in previous works \cite{stanford}. We trained two ResNet-34 baselines for the ECG and PPG signals. The Deep Mut$.$ baseline contains two separate models for the two signal modalities. The Stanford DeepBeat baseline only operates on PPG signals. We will use both AUROC (area under the receiver-operator-characteristic curve) and AUPRC (area under the precision-recall curve) as evaluation metrics. We also conducted bootstrapping tests with 1000 bootstrapping samples for statistical significance comparisons. 

\begin{figure*}[hbt!]
    \begin{subfigure}[b]{\textwidth}
        \centering
        \includegraphics[width=\textwidth]{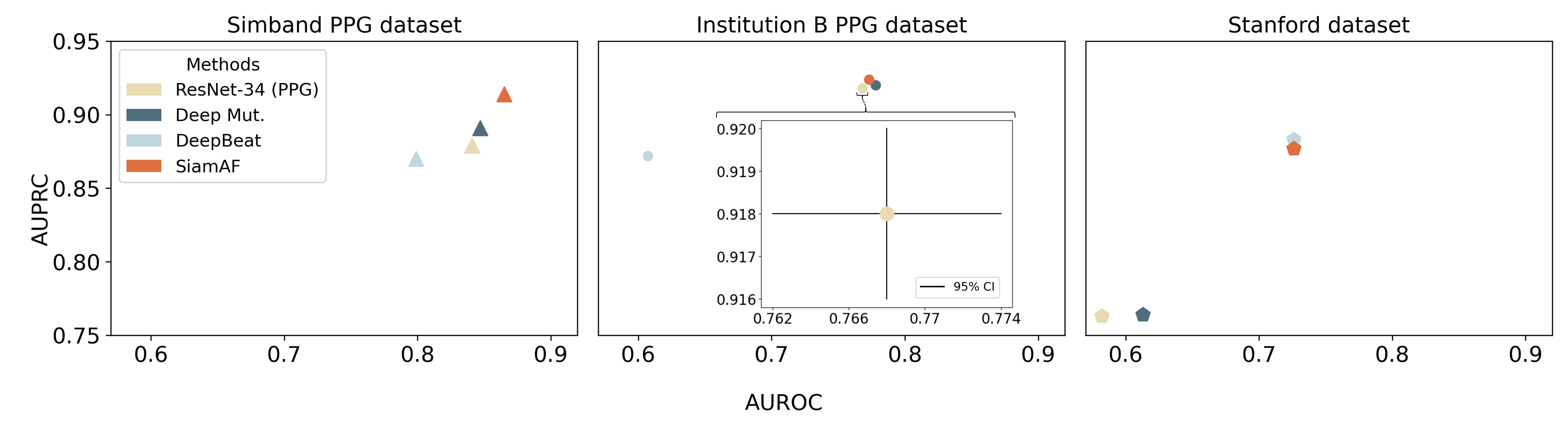}
        \caption{Performance comparisons of models on different PPG test sets.}
        \label{fig:ppgcomps}
    \end{subfigure}
    \newline
    \begin{subfigure}[b]{\textwidth}
        \centering
        \includegraphics[width=0.6667\textwidth]{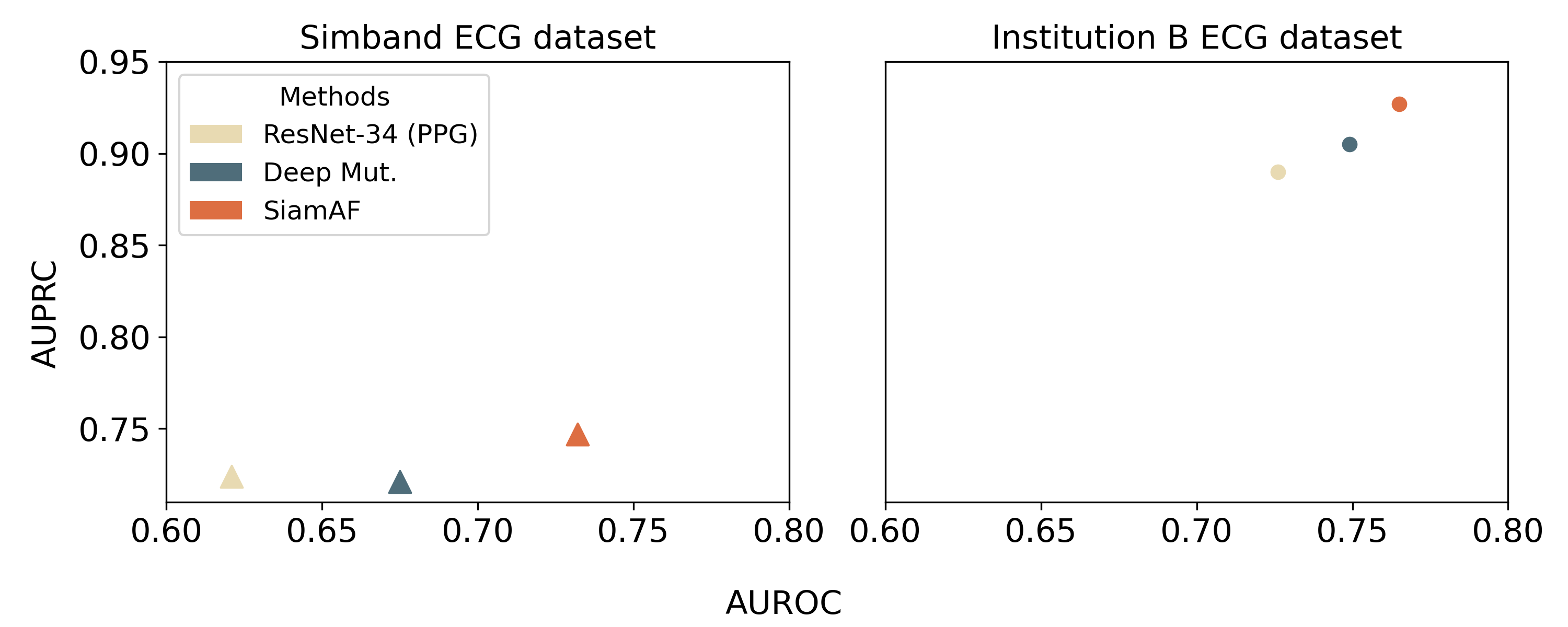}
        \caption{Performance comparisons of models on different ECG test sets.}
        \label{fig:ecgcomps}
    \end{subfigure}
    \caption{This figure shows the performance comparisons of all models on the test sets. Note the Stanford dataset does not contain ECG signals. In both plots, the horizontal axis is the AUPRC score, and the vertical axis is the AUROC score. The closer to the top right corner, the better. Here confidence intervals are generally smaller than the dot sizes, so we provide a zoom-in of one of the dots to demonstrate the 95\% CI. For exact CI and statistical significance test results, please refer to Sec.\ref{sec:perfvals} and Sec.\ref{sec:stattests}.}
    \label{fig:ppgecgcomps}
\end{figure*}

The evaluation results are shown in Figure \ref{fig:ppgecgcomps}. In the PPG test sets, the proposed method SiamAF performs comparably or substantially better than the baseline methods; \textbf{in the ECG test sets, our method outperforms the baseline methods by a significant margin}. For detailed values and the description of the bootstrapping calculation, please refer to Section \ref{sec:bootstrap}, Tables \ref{tab:appdxAUROC} and \ref{tab:appdxAUPRC} in the appendix.

\begin{figure*}[hbt!]
    \begin{subfigure}[b]{0.333\textwidth}
        \centering
        \includegraphics[width=\textwidth]{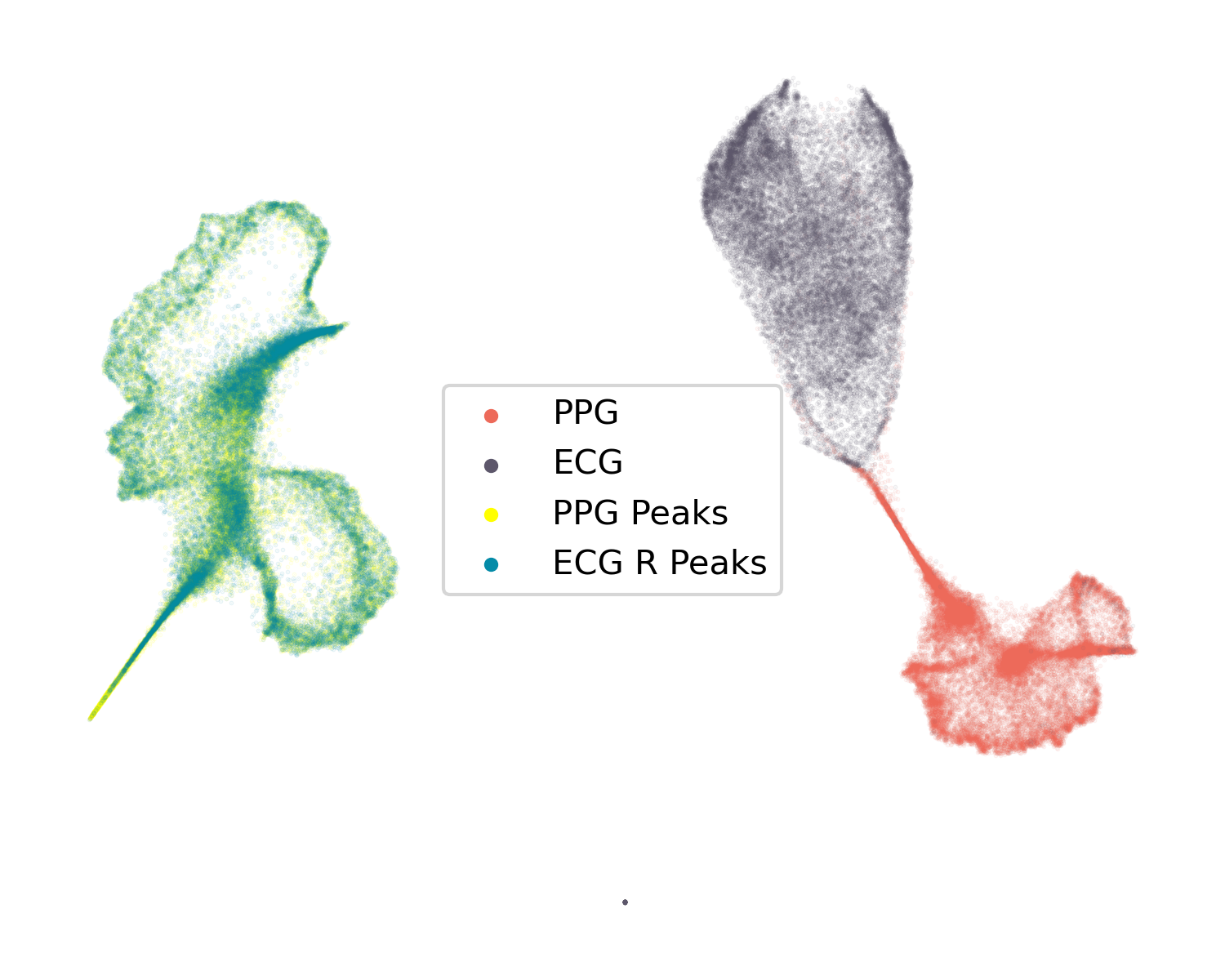}
        \caption{The first stage feature visualization.}
        \label{fig:simsiamlatent1}
    \end{subfigure}
    \begin{subfigure}[b]{0.333\textwidth}
        \centering
        \includegraphics[width=\textwidth]{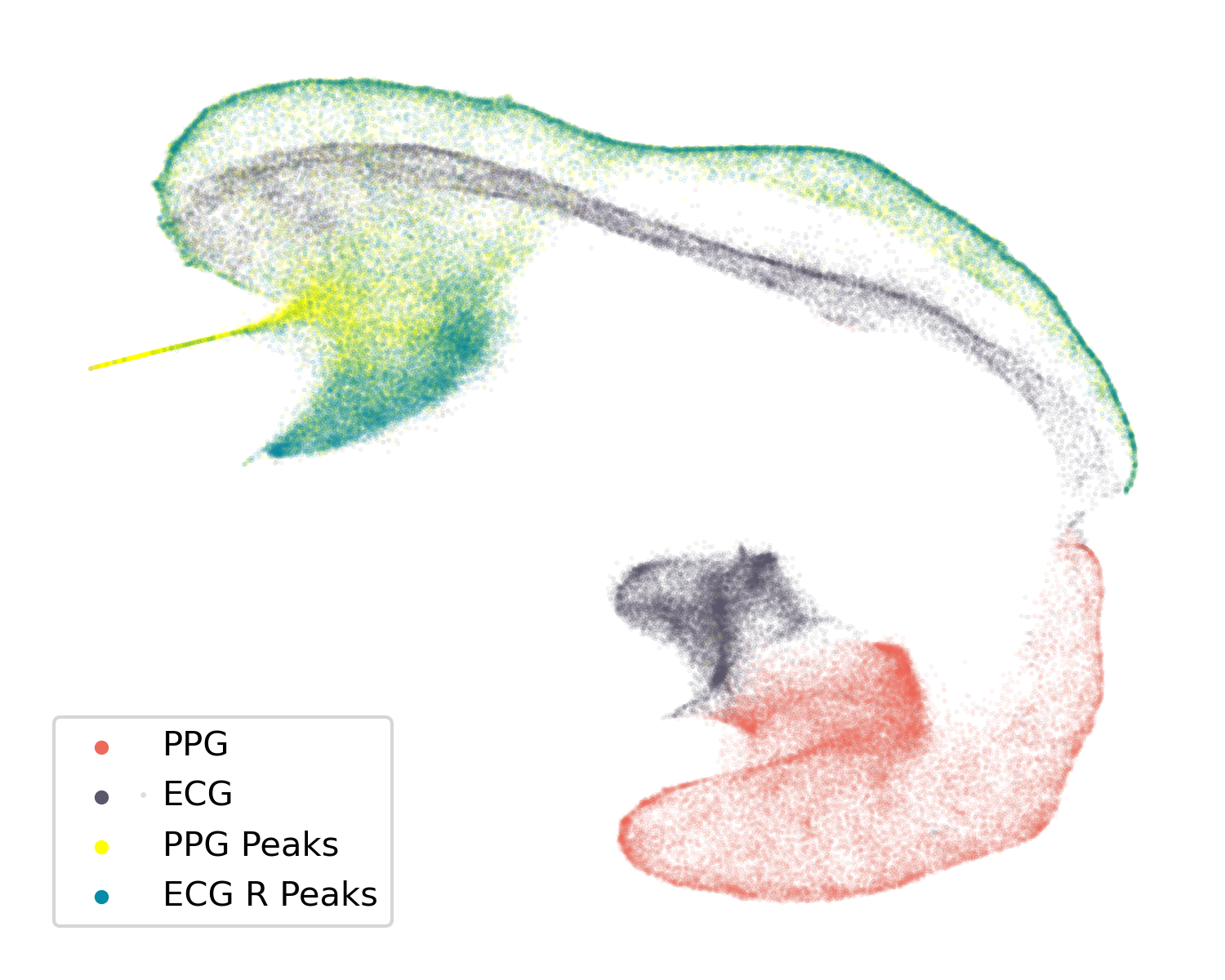}
        \caption{The second stage feature visualization.}
        \label{fig:simsiamlatent2}
    \end{subfigure}
    \begin{subfigure}[b]{0.333\textwidth}
        \centering
        \includegraphics[width=\textwidth]{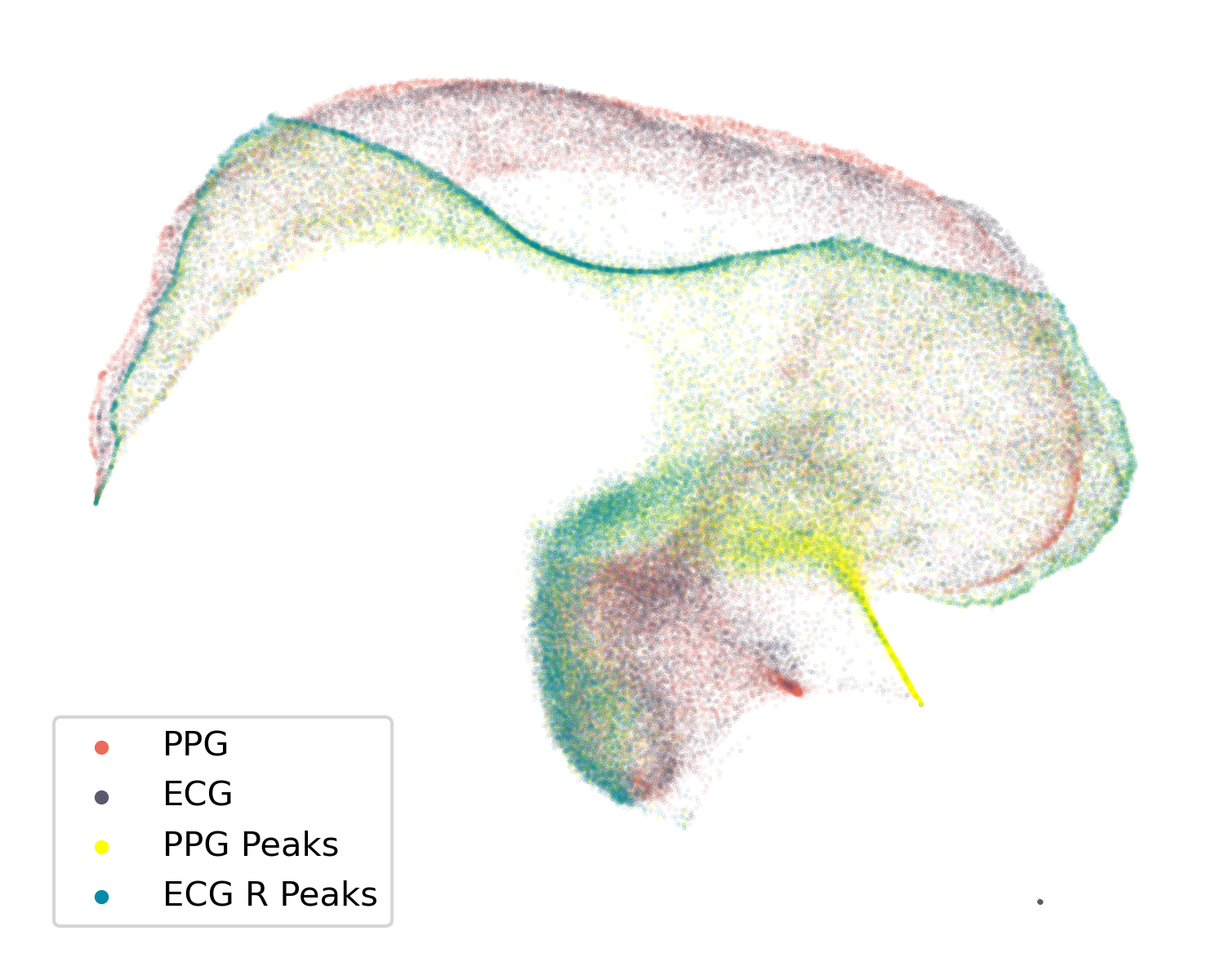}
        \caption{The third stage feature visualization.}
        \label{fig:simsiamlatent3}
    \end{subfigure}
    \newline
    \begin{subfigure}[b]{0.333\textwidth}
        \centering
        \includegraphics[width=\textwidth]{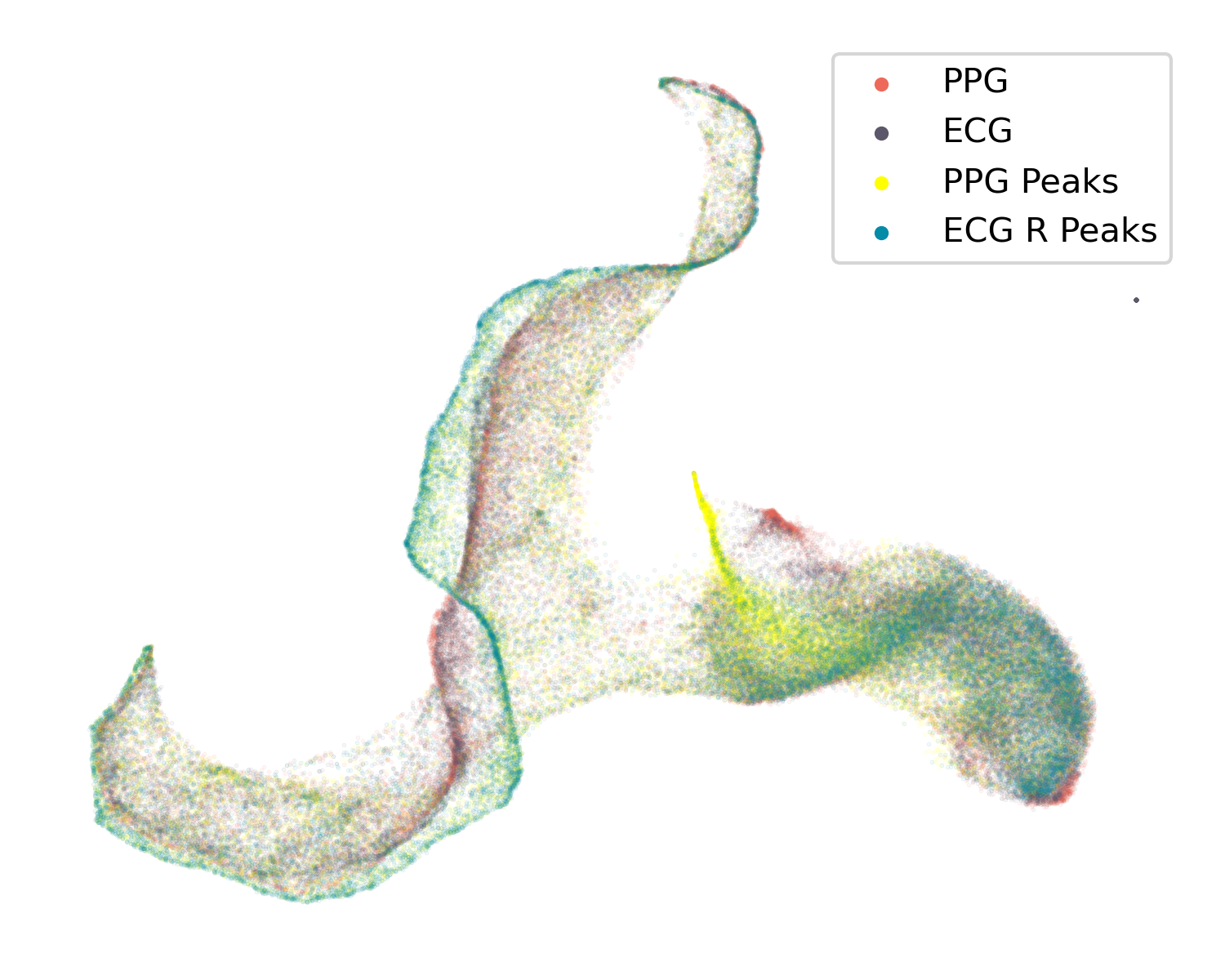}
        \caption{The fourth stage feature visualization.}
        \label{fig:simsiamlatent4}
    \end{subfigure}
    \begin{subfigure}[b]{0.666\textwidth}
        \centering
        \includegraphics[width=\textwidth]{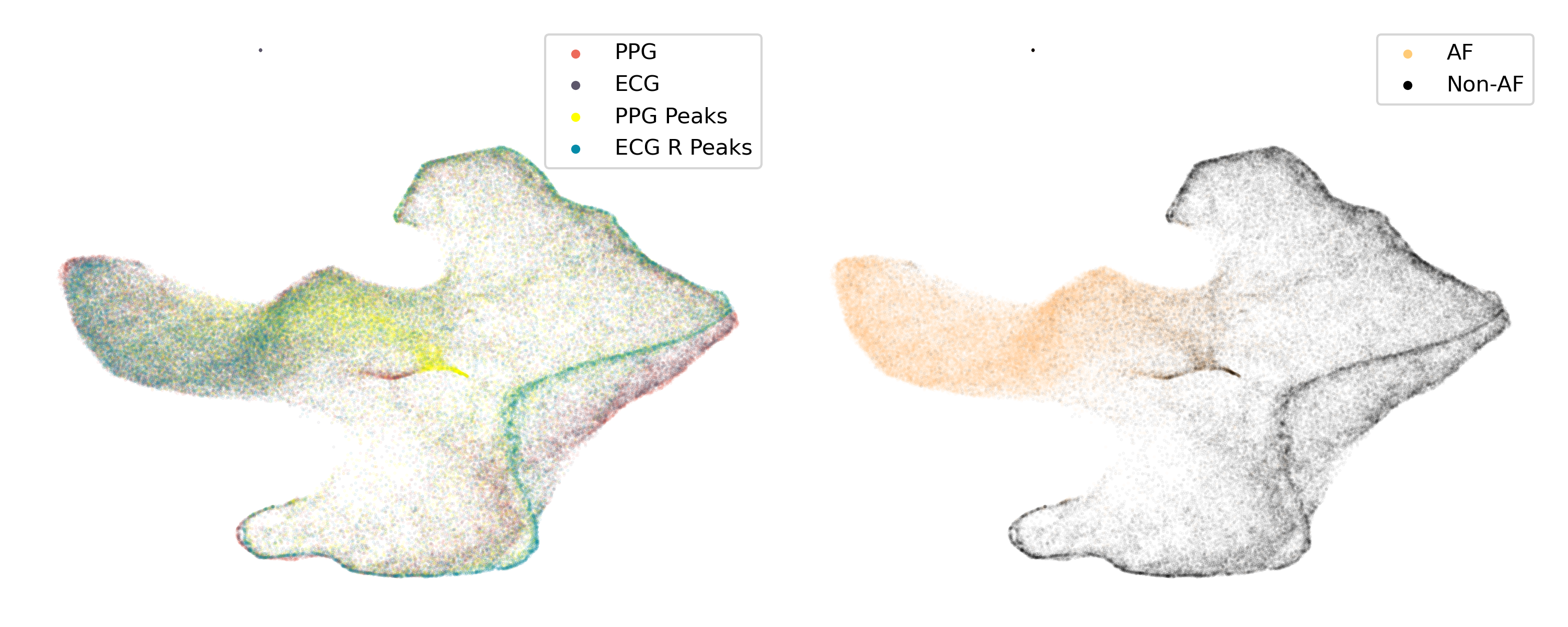}
        \caption{Final stage feature visualization.}
        \label{fig:simsiamlatent5}
    \end{subfigure}
    \caption{The four plots visualize latent features after each pooling layer of the ResNet-34 encoder used in our proposed framework. Figure \hyperref[fig:simsiamlatent1]{(a)}, \hyperref[fig:simsiamlatent2]{(b)}, \hyperref[fig:simsiamlatent3]{(c)}, \hyperref[fig:simsiamlatent4]{(d)}, \hyperref[fig:simsiamlatent5]{(e)}, each represent one of the stages. In Figure \hyperref[fig:simsiamlatent5]{(e)}, we also visualize the separation between AF and non-AF classes in the same latent space.}
    \label{fig:simsiamlatent}
\end{figure*}

\begin{figure}[hbt!]
    \centering
    \begin{subfigure}[b]{0.5\textwidth}
        \centering
        \includegraphics[width=\textwidth]{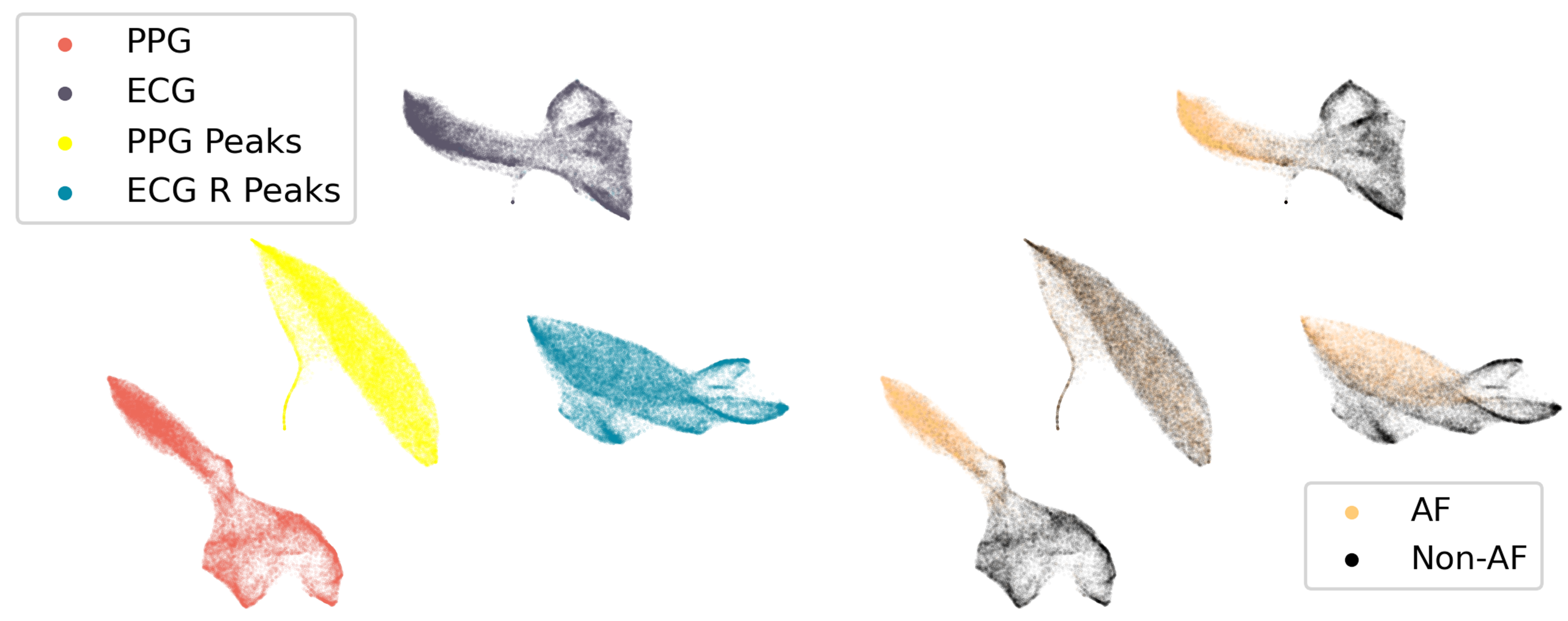}
        \caption{Final stage feature visualization of the ResNet-34 (ECG encoder and PPG encoders).}
    \end{subfigure}
    \newline
    \begin{subfigure}[b]{0.5\textwidth}
        \centering
        \includegraphics[width=\textwidth]{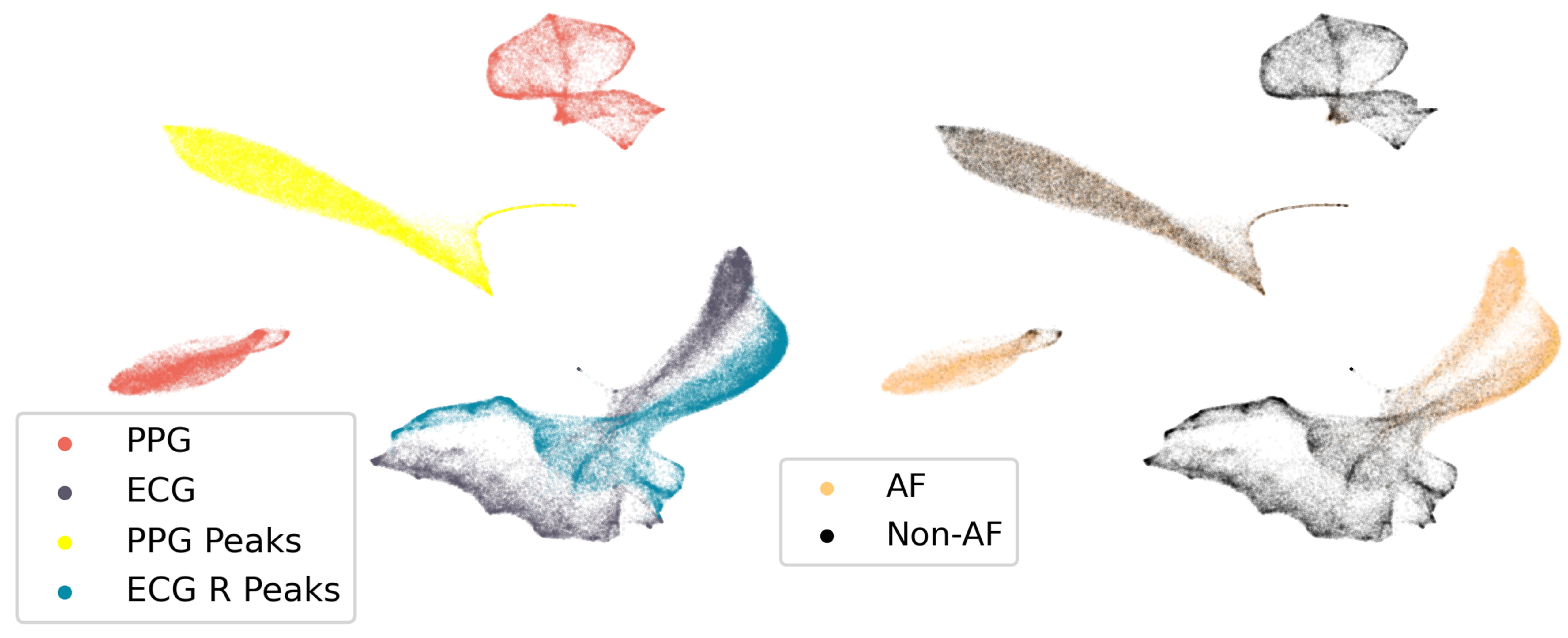}
        \caption{Final stage feature visualization of the Deep Mut$.$ encoders.}
    \end{subfigure}
    \caption{This figure shows visualizations of latent features at the final average pooling layer of the encoders in the baseline models.}
    \label{fig:baselinelatent}
\end{figure}



\begin{figure}[hbt!]
    \begin{subfigure}[b]{0.5\textwidth}
        \centering
        \includegraphics[width=\textwidth]{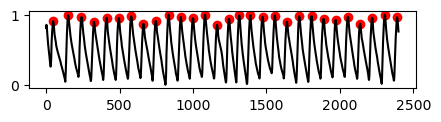}
        \caption{PPG signal peak detection example.}
    \end{subfigure}
    \begin{subfigure}[b]{0.5\textwidth}
        \centering
        \includegraphics[width=\textwidth]{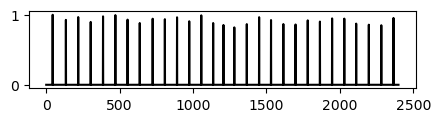}
        \caption{PPG peak only signal example.}
    \end{subfigure}
    \newline
    \begin{subfigure}[b]{0.5\textwidth}
        \centering
        \includegraphics[width=\textwidth]{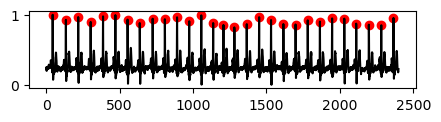}
        \caption{ECG signal R peak detection example.}
    \end{subfigure}
    \begin{subfigure}[b]{0.5\textwidth}
        \centering
        \includegraphics[width=\textwidth]{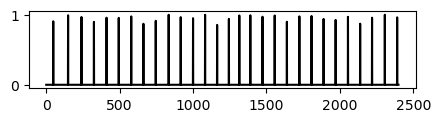}
        \caption{ECG peak only signal example.}
    \end{subfigure}
    \caption{This figure shows an example of the peak detection results on a pair of time-synchronized (simultaneous) 30s ECG and PPG signals and the corresponding encoded peak sequences. Here the red dots indicate the detected peaks. }
    \label{fig:peaks}
\end{figure}

\subsection{Exploration on learned representations and interpretability}
\label{sec: explorelatent}
ECG and PPG signals contain similar data, including RR (heartbeat) intervals and peaks, which are important diagnostic indicators for AF by human experts. We designed our model based on these facts, by encouraging it to learn and exploit the shared information between the two modalities for AF detection. Ideally, our model should behave similarly to humans utilizing this shared information (e.g., RR intervals) for the AF detection task. To verify our design approach, we explore the learned representation of the encoder in our proposed framework through dimension reduction and visualization of the activations from layers in the network. For visualization purposes, we randomly sub-sampled 1\% of the validation set and passed the samples through the learned ResNet-34 encoder in our proposed method. 

Medically, signal peaks and peak intervals from the PPG and ECG signals are crucial information for AF detection. 
Naturally, in a robust model, peak information should likewise contribute substantially toward AF classification.
For non-noisy data, the peaks are time-synchronized between PPG and ECG signals. By using only the peaks from the PPG signal, we can capture essential information for AF detection that is comparable to the information obtained from using the entire PPG signal. Likewise, the model should behave similarly when we retain only the peaks in ECG signals to when we retain the full ECG signal. Finally, the PPG and ECG signals contain similar information to each other, which means that for AF detection, they likely will map to the same neighborhood in the latent space.

We can verify these expected behaviors through visualization of the latent space.
To do so, we extract peak information from the PPG signals using the heartpy toolkit \cite{vanGent2019,vanGentHeartpy2019} and the R peaks in the ECG signals with NeuroKit2 \cite{neurokit2}. We create new peak-only PPG signals that are set to 0 everywhere except at the peaks and we do the same for ECG, preserving only all R-peaks. Figure \ref{fig:peaks} shows an example of the PPG detection and the peak-only signals. 
These peak-only signals allow us to feed the model only peak information and remove other waveform information. Thus we aim to verify whether (1) the peak-only PPG signals map to the same location in latent space as full PPG signals (and similarly for EEG) and (2) whether the PPG and EEG signals from the same time period of the same patient map to nearby locations in latent space.

We use PaCMAP \cite{pacmap} to visualize the feature outputs of each pooling layer, shown in Figure \ref{fig:simsiamlatent}. We also visualized feature outputs for both ECG and PPG-only and the Deep Mut$.$ baseline models in Figure \ref{fig:baselinelatent}. It is obvious that \textbf{at the later stage of the learned encoder in our proposed framework, both the PPG and ECG peak-only-signal features overlap almost perfectly with the original PPG and ECG-signal features}. This verifies hypotheses (1) and (2) above. In contrast, Figure \ref{fig:baselinelatent} shows that for the ResNet-34 and Deep Mut$.$ approaches, the encoded peak-only features never overlap with the PPG and ECG features, even at the final stages of the models. 

\subsection{Extension to semi-supervised learning with limited training labels}
\begin{figure*}[hbt!]
    \begin{subfigure}[b]{\textwidth}
        \centering
        \includegraphics[width=\textwidth]{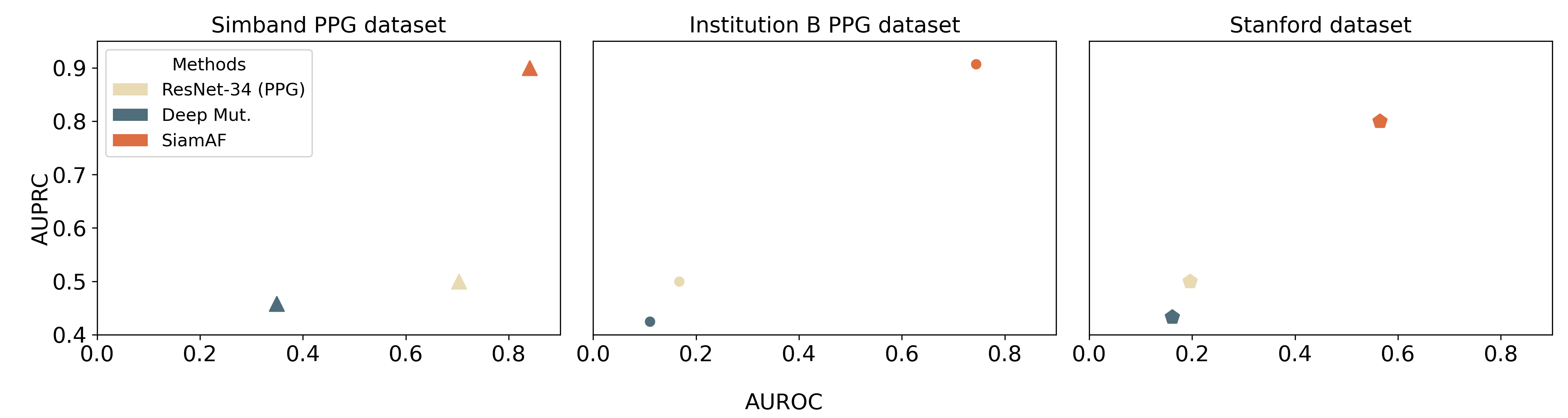}
        \caption{Performance comparisons of models trained with 1\% of the training set labels on different PPG test sets.}
        \label{fig:ppgcomps1perc}
    \end{subfigure}
    \newline
    \begin{subfigure}[b]{\textwidth}
        \centering
        \includegraphics[width=0.6667\textwidth]{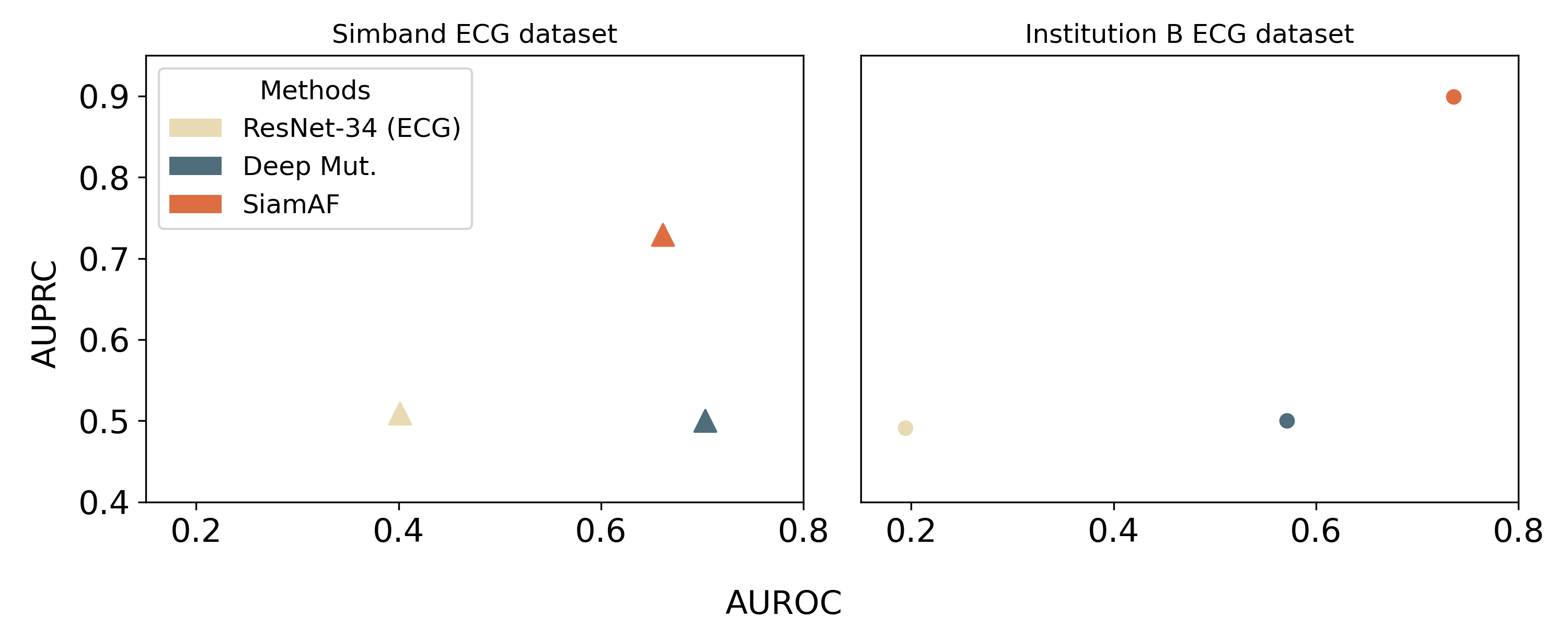}
        \caption{Performance comparisons of models trained with 1\% of the training set labels on different ECG test sets.}
        \label{fig:ecgcomps1perc}
    \end{subfigure}
    \caption{This figure shows the performance comparisons of all models on the test sets. The models evaluated in this experiment are all trained with 1\% of the training set labels. The figure arrangements are similar to that of Fig.\ref{fig:ppgcomps}. Again, confidence intervals are smaller than dot sizes, and thus are not displayed/visible.}
    \label{fig:ppgecgcomps1perc}
\end{figure*}

Here, we adapt our framework to the case where we have our usual training set, but only a small fraction of the data are labeled. We use the Siamese network to ensure that unlabeled samples can contribute to the learning process by constraining their PPG and ECG features to map to the same locations in latent space. This constraint is applied through learning the $L_{\agree}$ we defined in Eq.\ref{eq:agree}. Being able to learn from small labeled training sets is valuable since manual annotation is expensive. While we use auto-generated labels for our study, such labels may not be available for other tasks and applications. Thus, we eliminated 99\% of the labels. The changes to the algorithm are minimal, 
we only need to ensure that during gradient descent, each mini-batch has at least some labeled training data; the easiest way to do this is to duplicate the 1\% of labeled samples to use within each mini-batch. To compare with baseline methods, we trained them with 1\% of the labeled samples using normal mini-batch training. As shown in Figure \ref{fig:ppgecgcomps1perc}, the Deep Mut$.$ and ResNet-34 (PPG) baseline failed to learn or achieve meaningful predictive performance, with AUC near 0.5 (random guessing). In contrast, the proposed method achieved reasonable performance; in fact, it achieved comparable performance to that of the ResNet-34 baseline trained with the fully labeled training set. (Recall that ResNet-34 performed worse than our method when both were trained on the fully labeled dataset). For detailed values, please refer to Tables \ref{tab:appdxAUROC1perc} and \ref{tab:appdxAUPRC1perc} in the appendix.


\subsection{Hyper-parameter selection}
During training processes for all experiments, we used 30 training epochs; ResNet-34 baseline models used Adam as the optimizer and a learning rate of 0.0001; the proposed method was trained using the SGD optimizer, with the learning rate set to 0.1 and momentum set to 0.9. For the Deep Mut$.$ baseline, the weight of the KL-divergence loss was set to 0.9. For our proposed method, the $\lambda$ was set to 1. We also tested the effect of the hyper-parameter $\lambda$'s value in our proposed method; test performances were insensitive to the $\lambda$ value within the range of 0.2 to 1.2, in this study it was set to 1.0 by default.


\section{Discussion}


This paper provides a new framework to learn from simultaneous ECG and PPG signals for AF detection. The framework has several important benefits over other approaches: (1) it is more accurate for detecting AF; (2) its latent space reflects that it uses medically-relevant information for AF detection (namely the signal peaks); (3) it can detect AF from either ECG or PPG modalities using a single model; (4) it can be trained in a semi-supervised fashion, where using only 1\% of the labels, it can attain performance as good as the baseline approaches that use 100\% of the labels.

We make some observations beyond our direct results. First, it seems that learning from simultaneous PPG and ECG is better than learning from them asynchronously (or from either one alone). This can be seen because our method and the only baseline that uses simultaneous PPG/ECG (namely Deep Mut$.$) perform better than the other methods. 

Second, there is a large gap in performance between Deep Mut$.$ and our proposed approach. Since they both use simultaneous PPG/ECG, this gap is likely due to our proposed agreement loss, which directly encourages the encoder to learn coherent latent features between the two signal modalities. On the other hand, the Deep Mut$.$ setup relies on minimizing the KL divergence between the two prediction probability distributions, which is a weaker target that does not incentivize the encoders to focus on shared characteristics that exist in both the ECG and PPG signals. 
In fact, the two encoders in the two branches of Deep Mut$.$ could be vastly different and focus on less important features. 

Third, The DeepBeat baseline performed worse on both Simband and Institution B datasets compared to other baseline methods; this indicates the poor generalizability of the DeepBeat model to finger-tip recorded PPG signals in the Institution B dataset and out-of-distribution data in the Simband dataset. That is, DeepBeat does not seem to generalize beyond the type of data it was trained on, indicating that it is not focusing on the correct aspects of the signal for generalization. It was trained on the Stanford training set, and performs well on the Stanford test set, but not on other datasets. 
In contrast, our proposed method is trained on data collected in a hospital, yet attained comparable performance to DeepBeat on the Stanford test set, which demonstrates our strong generalizability to ambulatory data.

Fourth, we consider the question of how much non-peak information might be valuable for AF detection. As we saw, other methods seem to be using information other than peaks to detect AF. And, as we also saw, that information does not seem to generalize across modalities (PPG/ECG) and across data collection modes (ambulatory vs$.$ hospital setting). By experimenting with peak-only signals, it appears that, at least in our approach, largely the peak information is used; the model determined this was the important shared information between ECG and PPG signals to retain.

Fifth, we discuss the value and need for labels compared to the value of the agreement loss. Large labeled datasets are expensive to obtain, particularly for medical applications. Our approach performed as well as some of the baselines, even using only 1\% of the training labels, meaning that the agreement loss provided quite a lot of information to steer the learning algorithms toward the correct information. 

\section{Limitations and future works}
While it is useful to have training sets that contain noisy signals, in our case, the labels are also noisy, which could hurt the performance of the model. We may need to investigate a robust learning algorithm in future work to improve model performance. 

In the experiment in Sec.\ref{sec: explorelatent}, we showed that the projection of the peak-only-signal feature highly overlaps with ECG and PPG signal projections. However, it is not a complete overlap, thus we hypothesize that the model still leverages other potential morphological features (e.g., the presence of the P-wave in ECG) for AF classification. Future work should also focus on developing more interpretable networks, which can provide explicit explanations for information used for classification.

In future work, we may consider examining other heart conditions than AF. AF is a prominent heart condition, but many other tasks are worth investigating. It is possible that our framework will assist with those tasks as well. Future studies should also expand the signal modalities beyond ECG and PPG. In this study, ECG and PPG could be framed as augmented measurements of the original heartbeat, augmented through the human body and the collection process. Another direction for future work is data augmentation. As usual, if we knew the transformations of the data that are invariant to AF detection, we could build those into our data, yielding more robust AF detection. 

\section{Conclusion}
In this study, we proposed a new approach for AF detection, SiamAF. The proposed approach leverages a Siamese network architecture and a novel loss function, simultaneously learning from both ECG and PPG signal modalities. By learning common features between the two modalities (namely peaks), our proposed model achieves superior AF detection performance compared to previous studies. Our findings offer a promising avenue for improving AF detection and could potentially benefit patients with known AF, and patients who did not previously know they had AF.

\section{Acknowledgements}
The present work is partially supported by NIH grants R01HL166233, K23HL125251, and R01HL155711.

\section{Reproducibility}
All development code for this study is available publicly at \url{https://github.com/chengstark/SiamAF}. The retrospective use of the Institution A dataset in this study is conducted according to the terms of a data use agreement between UCSF and Emory University. As a result, the dataset is not available for public use without additional institutional approvals. Similarly, the Institution B dataset is not available for public use. To obtain access to the private datasets utilized in this study, please contact the corresponding author. Requests will undergo evaluation in accordance with the data use agreement established between institutions and may necessitate escalation to the relevant institutional data governance committee(s). The remaining external datasets used for evaluation in this study are publicly available.

\bibliographystyle{unsrt}  
\bibliography{main}  

\newpage
\section{Appendix}

\subsection{Bootstrap by patient calculations}
\label{sec:bootstrap}
We conducted bootstrapping tests with $N=1000$ samples to calculate the confidence interval for statistical significance comparisons. Each bootstrap sample is constructed by randomly sampling patients with replacements from each test set (i.e., $P$ random patients from each test set, where the test set contains $P$ unique patients). For each bootstrap sample, AUROC and AUPRC scores were calculated and collected. We calculated the 95\% confidence interval (CI) using $\mu \pm (1.96)\frac{\sigma}{\sqrt{N}}$; here $\mu$ is the mean of the collected bootstrap samples' AUPRC or AUROC scores, $\sigma$ is the standard deviation of the collected bootstrap samples' AUPRC or AUROC scores, and $N = 1000$ is the number of bootstrap samples. 

\subsection{Statistical significance test results}
\label{sec:stattests}
\justifying
This section contains the p-values of the pairwise statistical significance tests between SiamAF and other tested models. For comparing the AUROC scores between any two models, we used the pairwise Delong test \cite{delong1988comparing} and the Python FastDelong package \cite{fastdelong}. For comparing the AUPRC scores between any two models, we used the pairwise t-test. After adjusting the threshold $p$ value with Bonferroni correction ($p_{\textit{thresh}} = \frac{0.05}{\textit{\# of tests}}$), all pairwise comparisons are statistically significant except when comparing the proposed method and DeepBeat's AUPRC scores on the Standford dataset -- in this case, our proposed model performed comparably to the DeepBeat baseline on DeepBeat's \textit{in-distribution} test set. The tables below show the p-value results. 

\begin{table}[hbt!]
\centering
\begin{tabular}{|ccc|}
\hline
\multicolumn{1}{|c|}{\textbf{Compared Model}} & \multicolumn{1}{c|}{\textbf{$p_{AUROC}$}} & \textbf{$p_{AUROC}$} \\ \hline
\multicolumn{3}{|c|}{Simband PPG dataset}                                                                        \\ \hline
\multicolumn{1}{|c|}{ResNet-34(PPG)}          & \multicolumn{1}{c|}{0.00}                 & 0.00                 \\ \hline
\multicolumn{1}{|c|}{Deep Mut.}               & \multicolumn{1}{c|}{0.00}                 & 0.00                 \\ \hline
\multicolumn{1}{|c|}{DeepBeat}                & \multicolumn{1}{c|}{0.00}                 & 0.00                 \\ \hline
\multicolumn{3}{|c|}{Institution B PPG dataset}                                                                  \\ \hline
\multicolumn{1}{|c|}{ResNet-34(PPG)}          & \multicolumn{1}{c|}{0.00}                 & 0.00                 \\ \hline
\multicolumn{1}{|c|}{Deep Mut.}               & \multicolumn{1}{c|}{0.00}                 & 0.00                 \\ \hline
\multicolumn{1}{|c|}{DeepBeat}                & \multicolumn{1}{c|}{0.00}                 & 0.00                 \\ \hline
\multicolumn{3}{|c|}{Stanford test set}                                                                           \\ \hline
\multicolumn{1}{|c|}{ResNet-34(PPG)}          & \multicolumn{1}{c|}{0.00}                 & 0.00                 \\ \hline
\multicolumn{1}{|c|}{Deep Mut.}               & \multicolumn{1}{c|}{0.00}                 & 0.00                 \\ \hline
\multicolumn{1}{|c|}{DeepBeat}                & \multicolumn{1}{c|}{0.00}                 & 0.0046               \\ \hline
\multicolumn{3}{|c|}{Simband ECG dataset}                                                                        \\ \hline
\multicolumn{1}{|c|}{ResNet-34(ECG)}          & \multicolumn{1}{c|}{0.00}                 & 0.00                 \\ \hline
\multicolumn{1}{|c|}{Deep Mut.}               & \multicolumn{1}{c|}{0.00}                 & 0.00                 \\ \hline
\multicolumn{3}{|c|}{Institution B ECG dataset}                                                                    \\ \hline
\multicolumn{1}{|c|}{ResNet-34(ECG)}          & \multicolumn{1}{c|}{0.00}                 & 0.00                 \\ \hline
\multicolumn{1}{|c|}{Deep Mut.}               & \multicolumn{1}{c|}{0.00}                 & 0.00                 \\ \hline
\end{tabular}
\caption{Pairwise $p_{\textit{AUROC}}$ and $p_{\textit{AUPRC}}$ values between SiamAF and all tested models.}
\end{table}

\begin{table}[hbt!]
\centering
\begin{tabular}{|ccc|}
\hline
\multicolumn{1}{|c|}{\textbf{Compared Model}} & \multicolumn{1}{c|}{\textbf{$p_{AUROC}$}} & \textbf{$p_{AUROC}$} \\ \hline
\multicolumn{3}{|c|}{Simband PPG dataset}                                                                        \\ \hline
\multicolumn{1}{|c|}{ResNet-34(PPG)}          & \multicolumn{1}{c|}{0.00}                 & 0.00                 \\ \hline
\multicolumn{1}{|c|}{Deep Mut.}               & \multicolumn{1}{c|}{0.00}                 & 0.00                 \\ \hline
\multicolumn{3}{|c|}{Institution B PPG dataset}                                                                  \\ \hline
\multicolumn{1}{|c|}{ResNet-34(PPG)}          & \multicolumn{1}{c|}{0.00}                 & 0.00                 \\ \hline
\multicolumn{1}{|c|}{Deep Mut.}               & \multicolumn{1}{c|}{0.00}                 & 0.00                 \\ \hline
\multicolumn{3}{|c|}{Stanford test set}                                                                           \\ \hline
\multicolumn{1}{|c|}{ResNet-34(PPG)}          & \multicolumn{1}{c|}{0.00}                 & 0.00                 \\ \hline
\multicolumn{1}{|c|}{Deep Mut.}               & \multicolumn{1}{c|}{0.00}                 & 0.00                 \\ \hline
\multicolumn{3}{|c|}{Simband ECG dataset}                                                                        \\ \hline
\multicolumn{1}{|c|}{ResNet-34(ECG)}          & \multicolumn{1}{c|}{0.00}                 & 0.00                 \\ \hline
\multicolumn{1}{|c|}{Deep Mut.}               & \multicolumn{1}{c|}{0.00}                 & 0.00                 \\ \hline
\multicolumn{3}{|c|}{Institution B ECG dataset}                                                                  \\ \hline
\multicolumn{1}{|c|}{ResNet-34(ECG)}          & \multicolumn{1}{c|}{0.00}                 & 0.00                 \\ \hline
\multicolumn{1}{|c|}{Deep Mut.}               & \multicolumn{1}{c|}{0.00}                 & 0.00                 \\ \hline
\end{tabular}
\caption{Pairwise $p_{\textit{AUROC}}$ and $p_{\textit{AUPRC}}$ values between SiamAF and all tested models trained \textit{with 1\% training labels}.}
\end{table}


    
    
\subsection{Dataset participants demographic information}
In this section, we display the demographic information of the datasets used in this study in detail.
\begin{table}[ht!]
\centering
\begin{adjustbox}{max width=0.5\textwidth}
\begin{tabular}{|l|r|r|}
\hline
\textbf{Demographic Category} & \textbf{Patient Counts} & \textbf{Percentage} \\ \hline
\multicolumn{3}{|l|}{\textbf{Gender}} \\ \hline
Male & 15,330 & 53.7\% \\ \hline
Female & 13,203 & 46.2\% \\ \hline
Others & 6 & 0\% \\ \hline
\multicolumn{3}{|l|}{\textbf{Age}} \\ \hline
<22 years & 3,925 & 13.8\% \\ \hline
22-39 years & 2,715 & 9.5\% \\ \hline
40-54 years & 4,372 & 25.3\% \\ \hline
55-64 years & 5,370 & 18.8\% \\ \hline
$\geq$65 years & 12,157 & 42.6\% \\ \hline
\multicolumn{3}{|l|}{\textbf{Race}} \\ \hline
White or Caucasian & 15,890 & 55.7\% \\ \hline
Black or African American & 2,159  & 7.4\% \\ \hline
Asian & 4,364 & 15.0\% \\ \hline
Native Hawaiian or Other Pacific Islander & 426 & 1.46\% \\ \hline
American Indian or Alaska Native & 212 & 0.7\% \\ \hline
Unknown/Declined & 1,149  & 4\% \\ \hline
Others & 4,913 & 16.9\% \\ \hline
\end{tabular}
\end{adjustbox}
\caption{Patient demographic information in Institution A dataset}
\label{tab:ucsf_demographics}
\end{table}

\subsection{Performance comparisons and values}
\label{sec:perfvals}
This section includes the detailed evaluation results.

\centering
\begin{table*}[hbt!]
\begin{adjustbox}{max width=\textwidth}

\begin{tabular}{|l|c|c|c|c|c|}
\hline
Dataset/AUROC       & ResNet-34 (PPG) & ResNet-34 (ECG) & Deep Mut$.$ & DeepBeat                          & Proposed        \\ \hline
Simband (ECG)       & N/A              & 0.724 [0.722 0.727]          & 0.721 [0.718 0.723]    & N/A                & \textbf{0.747 [0.744 0.75]} \\ \hline
Simband (PPG)       & 0.879 [0.878 0.881]          & N/A              & 0.891 [0.889 0.892]     &0.870 [0.868 0.871]             & \textbf{0.914 [0.913 0.916]}  \\ \hline
Institution B (ECG) & N/A              & 0.89 [0.887 0.893]          & 0.905 [0.902 0.908]    & N/A                              & \textbf{0.927 [0.925 0.929]} \\ \hline
Institution B (PPG) & 0.918 [0.916 0.92]          & N/A              & 0.92 [0.918 0.922]    &0.872 [0.87 0.875]              & \textbf{0.924 [0.922 0.925]} \\ \hline
Stanford test set   & 0.763 [0.761 0.764]          & N/A              & 0.764 [0.763 0.766]    & \textbf{0.883 [0.882 0.884]} & 0.877 [0.876 0.878]          \\ \hline
\end{tabular}
\end{adjustbox}

\caption{AUROC comparisons of models on different test sets. N/A means that the method only works for the other modality (either ECG or PPG).}
\label{tab:appdxAUROC}
\end{table*}

\begin{table*}[hbt!]
\centering
\begin{adjustbox}{max width=\textwidth}

\begin{tabular}{|l|c|c|c|c|c|}
\hline
Dataset/AUPRC       & ResNet-34 (PPG) & ResNet-34 (ECG) & Deep Mut$.$       & DeepBeat                            & Proposed        \\ \hline
Simband (ECG)       & N/A              & 0.621 [0.617 0.625]          & 0.675 [0.671 0.679]          & N/A            & \textbf{0.732 [0.729 0.736]} \\ \hline
Simband (PPG)       & 0.841 [0.838 0.843]          & N/A              & 0.847 [0.844 0.845]          & 0.799 [0.796 0.803]   & \textbf{0.865 [0.863 0.868]} \\ \hline
Institution B (ECG) & N/A              & 0.726 [0.72 0.732]          & 0.749 [0.743 0.755]          & N/A           & \textbf{0.765 [0.759 0.772]} \\ \hline
Institution B (PPG) & 0.768 [0.762 0.774]          & N/A              & \textbf{0.778 [0.772 0.784]} & 0.607 [0.601 0.614]      & 0.773 [0.768 0.779]          \\ \hline
Stanford test set   & 0.582 [0.577 0.586]          & N/A              & 0.613 [0.609 0.617]           & \textbf{0.726 [0.723 0.73]}     & 0.726 [0.722 0.729]          \\ \hline
\end{tabular}
\end{adjustbox}
\caption{AUPRC comparisons of models on different test sets.}
\label{tab:appdxAUPRC}
\end{table*}

\begin{table*}[hbt!]
\centering
\begin{tabular}{|l|c|c|c|c|}
\hline
Dataset/AUROC       & ResNet-34 (PPG) & ResNet-34 (ECG) & Deep Mut$.$ & Proposed        \\ \hline
Simband (ECG)       & N/A              & 0.509 [0.509 0.51]            & 0.5 [0.5 0.5]       & \textbf{0.729 [0.726 0.732]} \\ \hline
Simband (PPG)       & 0.5 [0.5 0.5]             & N/A              & 0.458 [0.456 0.459]    & \textbf{0.9 [0.898 0.901]} \\ \hline
Institution B (ECG) & N/A              & 0.491 [0.49 0.492]             & 0.5 [0.5 0.5]       & \textbf{0.899 [0.897 0.902]} \\ \hline
Institution B (PPG) & 0.5 [0.5 0.5]          & N/A               & 0.425 [0.423 0.426]    & \textbf{0.907 [0.905 0.909]} \\ \hline
Stanford test set   & 0.5 [0.5 0.5]          & N/A               & 0.433 [0.432 0.434]    & \textbf{0.8 [0.799 0.801]} \\ \hline
\end{tabular}
\caption{AUROC comparisons of models on different test sets when trained with 1\% of the training labels.}
\label{tab:appdxAUROC1perc}
\end{table*}

\begin{table*}[hbt!]
\centering
\begin{tabular}{|l|c|c|c|c|}
\hline
Dataset/AUROC       & ResNet-34 (PPG) & ResNet-34 (ECG) & Deep Mut$.$ & Proposed        \\ \hline
Simband (ECG)       & N/A               & 0.401 [0.393 0.41]           & \textbf{0.703 [0.702 0.705]}    & 0.661 [0.658 0.665] \\ \hline
Simband (PPG)       & 0.703 [0.702 0.705]           & N/A               & 0.349 [0.345 0.352]    & \textbf{0.841 [0.838 0.844]} \\ \hline
Institution B (ECG) & N/A               & 0.194 [0.19 0.198]          &0.571 [0.57 0.572]    & \textbf{0.736 [0.731 0.742]} \\ \hline
Institution B (PPG) & 0.167 [0.164 0.171]          & N/A               & 0.11 [0.109 0.112]    & \textbf{0.744 [0.739 0.749]} \\ \hline
Stanford test set   & 0.196 [0.193 0.2]          & N/A               & 0.161 [0.159 0.163]    & \textbf{0.565 [0.56 0.569]} \\ \hline
\end{tabular}
\caption{AUPRC comparisons of models on different test sets when trained with 1\% of the training labels.}
\label{tab:appdxAUPRC1perc}
\end{table*}


\end{document}